\documentclass[10pt,twocolumn,letterpaper]{article}

\usepackage{cvpr}
\usepackage{times}
\usepackage{epsfig}
\usepackage{graphicx}
\usepackage{amsmath}
\usepackage{amssymb}
\usepackage{hhline}
\usepackage{graphics}
\usepackage{makecell}

\usepackage[pagebackref=true,breaklinks=true,letterpaper=true,colorlinks,bookmarks=false]{hyperref}

\cvprfinalcopy 

\def\cvprPaperID{2430} 

\ifcvprfinal\fi
\begin{document}

\title{Recurrent Attentional Networks for Saliency Detection}

\author{Jason Kuen, Zhenhua Wang, Gang Wang\thanks{Corresponding author}\\
School of Electrical and Electronic Engineering,\\
Nanyang Technological University.\\
{\tt\small \{jkuen001,wzh,wanggang\}@ntu.edu.sg}
}

\maketitle

\begin{abstract}
Convolutional-deconvolution networks can be adopted to perform end-to-end saliency detection. But, they do not work well with objects of multiple scales. To overcome such a limitation, in this work, we propose a recurrent attentional convolutional-deconvolution network (RACDNN). Using spatial transformer and recurrent network units, RACDNN is able to iteratively attend to selected image sub-regions to perform saliency refinement progressively. Besides tackling the scale problem, RACDNN can also learn context-aware features from past iterations to enhance saliency refinement in future iterations. Experiments on several challenging saliency detection datasets validate the effectiveness of RACDNN, and show that RACDNN outperforms state-of-the-art saliency detection methods.
\end{abstract}

\thispagestyle{empty}
\section{Introduction}

Saliency detection refers to the challenging computer vision task of identifying salient objects in imagery and segmenting their object boundaries. Despite that it has been studied for years, saliency detection still remains an unsolved research problem due to its tough goal to model high-level subjective human perceptions. Recently, saliency detection methods have received considerable amount of attention, as there is a wide and growing range of applications facilitated by it. Some of the notable applications of saliency detection are object recognition \cite{ren2014region}, visual tracking \cite{borji2012adaptive}, and image retrieval \cite{chen2009sketch2photo}.

Traditionally, methods in saliency detection leverage low-level saliency priors such as contrast prior and center prior to model and approximate human saliency. However, such low-level priors can hardly capture high-level information about the objects and its surroundings: the traditional methods are still very far away from how saliency works in the context of human perceptions. To incorporate high-level visual concepts into a saliency detection framework, it is natural to consider \textbf{convolutional neural networks (CNN)}. For a lot of computer vision tasks \cite{gu2015recent}, CNNs have shown to be remarkably effective. It is also the first learning algorithm to achieve human-competitive performances \cite{he2015delving} in large-scale image classification task, which is a high-level vision task like saliency detection. Although there have been works on developing CNNs for visual saliency modeling, they either focus on predicting eye fixations \cite{nian2015predicting}, or applying CNNs to predict just the saliency value of visual sub-units (e.g. superpixels) independently \cite{zhao2015saliency}. Besides, conventional CNNs downsize feature maps over multiple convolutional and pooling layers and lose detailed information for our problem of densely segmenting salient objects.

\begin{figure}[t]
	\centering
	\includegraphics[width=\columnwidth]{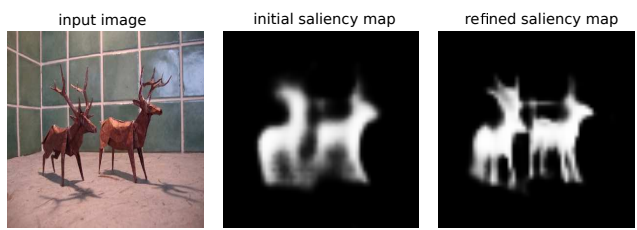}
	\caption{An example of applying recurrent attention-based saliency refinement to an initial saliency map produced by convolutional-deconvolutional network. Compared to the initial saliency map, the refined saliency map has significantly sharper edges and preserves more object details.}
	\label{fig:front_qualitative}
\end{figure}

Inspired by the success of \textbf{convolutional-deconvolutional network (CNN-DecNN)} in semantic segmentation \cite{noh2015learning}, in this paper, we adapt the network to detect salient objects in an end-to-end fashion. For this framework, the input is an image, and the output is its corresponding saliency map. A deconvolutional network (DecNN) is a variant of CNN that performs convolution and unpooling to produce dense pixel-precise outputs. However, CNN-DecNN works poorly for objects of multiple scales \cite{long2015fully,noh2015learning} due to the fixed-size receptive fields. To overcome this limitation, we propose a \textbf{recurrent attentional convolutional-deconvolutional network (RACDNN)} to refine the saliency maps generated by CNN-DeCNN. RACDNN uses spatial transformer and recurrent network units to iteratively attend to flexibly-sized image sub-regions, and refines the saliency predictions on those sub-regions. As shown in Figure \ref{fig:front_qualitative}, RACDNN can perform saliency detection at finer scales due to its ability to attend to smaller sub-regions. Another advantage of RACDNN is that the attended sub-regions in the previous iterations can provide contextual information for the saliency refinement of the sub-region in the current iteration. For example, in Figure \ref{fig:front_qualitative}, RACDNN can make use of the more visible front legs of the deers to help at refining the saliency values of the less-visible back legs. 

We perform experiments on several challenging saliency detection benchmark datasets, and compare the proposed method with state-of-the-art saliency detection methods. Experimental results show the effectiveness of our proposed method. 

\section{Related work}
Saliency detection methods can be coarsely categorized into bottom-up and top-down methods. Bottom-up methods \cite{itti1998model,harel2006graph,hou2007saliency,achanta2009frequency,liu2011learning,cheng2015global,margolin2013what} make use of level local visual cues like color, contrast, orientation and texture. Top-down methods \cite{zhang2008sun,yang2012top,judd2009learning} are based on high-level task-specific prior knowledge. Recently, deep learning-based saliency detection methods \cite{wang2015deep,zhang2015co,zhao2015saliency,li2015visual,vig2014large} have been very successful. Instead of manually defining and tuning saliency-specific features, these methods can learn both low-level features and high-level semantics useful for saliency detection straight from minimally processed images. However, these works employ neither attention mechanism nor RNN to improve saliency detection. To the best of our knowledge, ours is the first work to exploit recurrent attention along with deep learning for saliency detection.

Attention models are a new variant of neural networks aiming to model visual attention. They are often used with recurrent neural networks to achieve sequential attention. \cite{mnih2014recurrent} formulates a recurrent attention model that surpasses CNN on some image classification tasks. \cite{ba2015multiple} extends the work of \cite{mnih2014recurrent} by making the model deeper and apply it for multi-object classification task. To overcome the training difficulty of recurrent attention model, \cite{gregor2015draw} propose a differentiable attention mechanism and apply it for generative image generation and image classification. \cite{jaderberg2015spatial} propose a differentiable and efficient sampling-based spatial attention mechanism, in which any spatial transformation can be used. Unlike the above works \cite{mnih2014recurrent, ba2015multiple, gregor2015draw} which mostly use small attention networks for low-resolution digit classification task, the attention mechanism used in our work is much more complex, as it is tied with a large CNN-DecNN for dense pixelwise saliency refinement.

\section{Proposed Method}
In this section, we describe our proposed saliency detection method in detail. In our method, initial saliency maps are first generated by a convolutional-deconvolutional network (CNN-DecNN) which takes entire images as input, and outputs saliency maps. The saliency maps are then refined iteratively via another CNN-DecNN operated under a recurrent attentional framework. Unlike the initial saliency map prediction which is done through single feedforward passes on the entire images, the saliency refinement is done locally on selected image sub-regions in a progressive way. At every processing iteration, the recurrent CNN-DecNN attends to an image sub-region, through the use of a spatial transformer-based attention mechanism. The attentional saliency refinement helps to alleviate the inability of CNN-DecNN to deal with multiscale saliency detection. In addition, the sequential nature of the attention enables the network to exploit contextual patterns from past iterations to enhance the representation of the attended sub-region, hence to improve the saliency detection performance.

\begin{figure}[ht]
	\centering
	\includegraphics[width=\columnwidth]{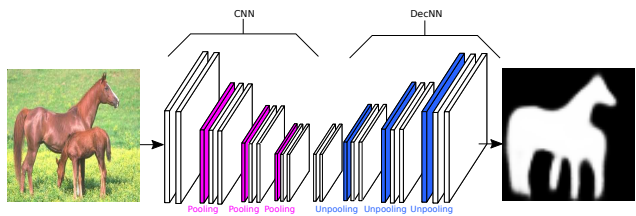}
	\caption{A generic convolutional-deconvolutional network for saliency detection.}
	\label{fig:cnn-decnn}
\end{figure}

\subsection{Deconvolutional Networks for Salient Object Detection}
Conventionally, CNNs downsize feature maps over multiple convolutional and pooling layers, to construct spatially compact image representations. Although these spatially compact feature maps are well-suited for whole-image classification tasks, they tend to produce very coarse outputs when being applied for dense pixelwise prediction tasks (e.g., semantic segmentation). To tackle dense prediction tasks in the multi-layered convolutional learning setting, one can append a deconvolutional network (DecNN) to a CNN as shown in \cite{noh2015learning}. In such a convolutional-deconvolutional (CNN-DecNN) framework, the CNN learns globally meaningful representations, while the DecNN upsizes feature maps and learns increasingly localized representations. Unlike the work of \cite{noh2015learning}, we preserve the spatial information of CNN's output (the input to DecNN) by using only convolutional layers. In practice, we find that preserving such spatial information works better than without preserving it. This is because the preserved spatial information provides a good head start for DecNN to gradually introduce more spatial information to the feature maps. A generic network architecture of CNN-DecNN is shown in Figure \ref{fig:cnn-decnn}.

A DecNN is almost identical to conventional CNNs except for a few minor differences. Firstly, in deconvolutional networks, convolution operations are often carried out in such a way that the resulting feature maps retain the same spatial sizes as those of the input feature maps. This is done by adding appropriate zero paddings beforehand. Secondly, the pooling operators adopted by CNNs are substituted with unpooling operators in DecNNs. Given input feature maps, unpooling operators work by upsizing the feature maps, contrary to what pooling operators achieve. A few variants of unpooling methods \cite{dosovitskiy2015learning,noh2015learning} have been proposed previously to tackle several computer vision tasks involving spatially large and dense outputs. In this paper, we employ the simple unpooling method demonstrated in \cite{dosovitskiy2015learning}, whereby each block (with spatial size $ 1\times1 $) in the input feature maps is mapped to the top left corner of a blank output block with spatial size $ k \times k $. This effectively increases the spatial size of the whole feature maps by a factor of $ k $. 

In the processing pipeline of CNN-DecNN for saliency detection, the CNN first transforms the input image $ x $ to a spatially compact hidden representation $ z $, as $ z = \mbox{\textit{CNN}}(x) $. Then, $ z $ is transformed to a raw saliency map $ r $ through the DecNN, as $ r = \mbox{\textit{DecNN}}(z) $. To obtain the final saliency map $ \bar{S} $ that lies within the probability range of $ [0,1] $, we perform $ \bar{S} = \sigma(r) $, passing the raw saliency map $ r $ into element-wise sigmoid activation function $ \sigma(\cdot) $. Given the groundtruth saliency map $ \bar{G} $, the loss function of CNN-DecNN for saliency detection is the binary cross-entropy between $ \bar{G} $ and $ \bar{S} $. The resulting network can be trained in end-to-end fashion to perform saliency detection. Although CNN-DecNN can achieve pixelwise labeling, it works poorly for objects of multiple scales \cite{long2015fully,noh2015learning} due to the fixed-size receptive fields used. Furthermore, long-distance contextual information which is important for saliency detection, cannot be well captured by the locally applied convolution filters in DecNN. To address these issues, we propose an recurrent attentional network that iteratively attends to image sub-regions (of unconstrained scale and location) for saliency refinement, which is described in the next two subsections.

\begin{figure}[ht]
\centering
	\includegraphics[width=130pt]{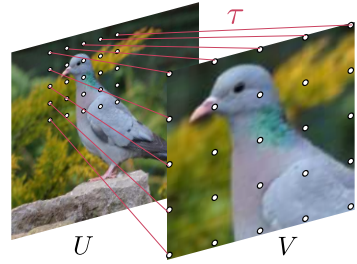}
	\caption{To map the input feature maps $ U $ to output feature maps $ V $, spatial tansformer transforms output point coordinates on $ V $ to sampling point coordinates on $ U $.}
	\label{fig:spatial_transformer}
\end{figure}

\subsection{Attentional Inputs and Outputs with Spatial Transformer}
To realize the attention mechanism for saliency refinement, we adopt the spatial transformer network proposed in \cite{jaderberg2015spatial}. Spatial transformer is a sub-differentiable sampling-based neural network which spatially transform its input feature maps (may also be images), resulting in an output feature maps that is an attended region of the input feature maps. Due to its differentiability, spatial transformer is relatively easier to train compared to some non-differentiable neural network-based attention mechanisms \cite{mnih2014recurrent,ba2015multiple} proposed recently.

Spatial transformer achieves spatial attention by mapping an input feature map $ U \in \mathbb{R}^{A \times B \times C} $ into an output feature map $ V \in \mathbb{R}^{A' \times B' \times C} $. $ V $ can have spatial sizes different from $ U $, but they must share the same number of channels $ C $ since we consider only spatial attention. Given $ U $, spatial transformer first computes the transformation matrix $ \tau $ that determines how the point coordinates in $ V $ are transformed to those in $ U $. An example of V-to-U coordinatewise transformation is shown in Figure \ref{fig:spatial_transformer}. A wide range of transformation types are supported by spatial transformer. For simplicity, we restrict the transformation to a basic form of spatial attention, involving only isotropic scaling and translation. The affine transformation matrix $ \tau $ with just isotropic scaling and translation is given as

\begin{equation} 
\tau = 
\begin{bmatrix}
a_s & 0 	& a_{tx} \\
0 	& a_s 	& a_{ty} \\
0	& 0		& 1
\end{bmatrix}
\end{equation}
where $ a_s $, $ a_{tx} $, and $ a_{ty} $ are the scaling, horizontal translation, and vertical translation parameters respectively. Aligning with the recent works \cite{mnih2014recurrent,ba2015multiple,gregor2015draw} in recurrent visual attention modeling, the parameters deciding where the attention takes place (in our case, $ \tau $) is produced by the localization network $ f_{loc}(\cdot) $. 
More details on $ f_{loc}(\cdot) $ will be introduced in Equation 9 in Section 3.3.  Subsequently, the transformation matrix $ \tau $ is applied to the regular coordinates of $ V $ to obtain sampling coordinates. Based on the sampling coordinates, $ V $ is formed by sampling feature map points from $ U $ using bilinear interpolation.

Generally, attention mechanisms are applied only to input images. However, our saliency refinement method (see Section 3.3) via DecNN demands that the input and output ends point to the same image sub-region. To this end, we propose an inverse spatial transformer which can map refined saliency output back to the same sub-region attended at input end. Assuming that $ \tau $ is the transformation matrix for the input end, the inverse spatial transformer takes the inverse of $ \tau $ as the output transformation matrix $ \tau^{-1} $:

\begin{equation} 
\tau^{-1} = 
\begin{bmatrix}
1/a_s 	& 0 		& -a_{tx}/a_s \\
0 		& 1/a_s 	& -a_{ty}/a_s \\
0		&			& 1
\end{bmatrix}
\end{equation}

\subsection{Recurrent Attentional Networks for Saliency Refinement}
Recurrent neural networks (RNN) \cite{elman1990finding} are a class of neural networks developed for modeling the sequential dependencies between sub-instances of sequential data. In RNN, the hidden state $ h_i $ at time step or iteration $ i $ is computed as a non-linear function of the input and the previous iteration's hidden state $ h_{i-1} $. Given an input $ x_i $ at iteration $ i $, the hidden state $ h_i $ of a RNN is formulated as:

\begin{equation}
h_i = \phi(W_Ix_i+ W_Rh_{i-1} + b)
\end{equation}
where $ W_I $ and $ W_R $ are the learnable weights for input-to-hidden and hidden-to-hidden connections respectively, while $ b $ is a bias term, and $ \phi(\cdot) $ is a nonlinear activation function. By explicitly making the current hidden state $ h_i $ dependable on the previous hidden state $ h_{i-1} $ , RNN is able to encode contextual information gained from past iterations for use in future iterations. As a result, a more powerful representation $ h_i $ can be learned.

\begin{figure}[ht]
\centering
	\includegraphics[width=\columnwidth]{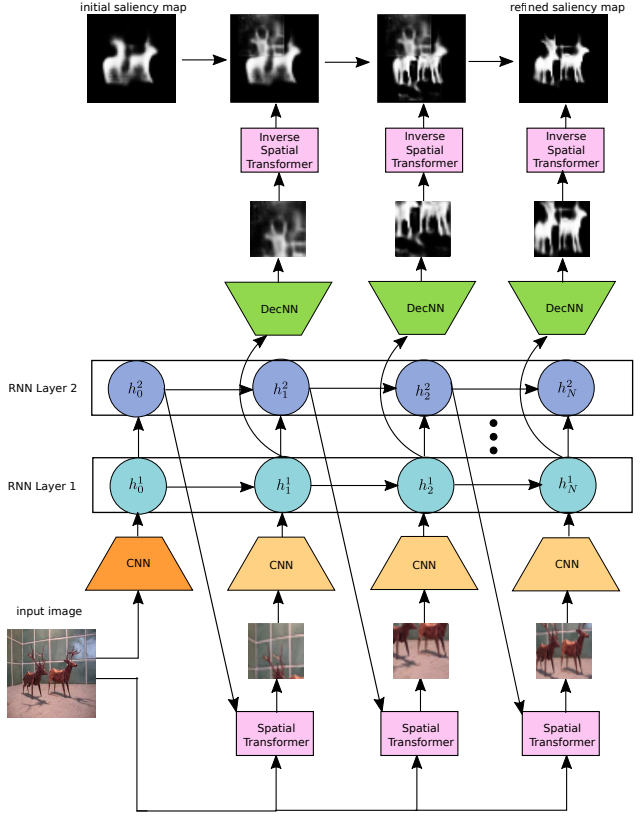}
	\caption{Overall architecture of our Recurrent Attentional Convolutional-Deconvolutional Network (RACDNN)}
	\label{fig:racdnn}
\end{figure}

In this work, we combine the recurrent computational structure of RNN with CNN-DecNN as well as the spatial transformer attention mechanism, to establish the \textbf{recurrent attentional convolutional-deconvolutional networks (RACDNN)}. As illustrated in Figure \ref{fig:racdnn}, given an intiail saliency map produced by the initial CNN-DeCNN, RACDNN iteratively uses spatial transformer to attend to a sub-region, and applies its CNN-DecNN to perform saliency refinement for the attended sub-region, by learning powerful context-aware features using RNN.

At every computational iteration $ i $, RACDNN first receives an attended input $ x_i $ from the full input image $ x $ as follows:

\begin{equation}
x_i = \mbox{\textit{ST}}(x,\tau_i)
\end{equation}
where \textit{ST}($ \cdot $) is a spatial transformer function which produces an output image sampled from the input image, given the transformation matrix $ \tau_i $. $ \tau_i $ is computed at the previous iteration $ i-1 $ through the localization network $ f_{loc}(\cdot) $. Then, RACDNN uses a recurrent-based CNN $ \mbox{\textit{CNN}}_r $ to encode the attended input $ x_i $ into a spatially-compact hidden representation $ z_i $. $ \mbox{\textit{CNN}}_r $ is similar to \textit{CNN} except that $ \mbox{\textit{CNN}}_r $ is used in the recurrent setting, and all recurrent instances of $ \mbox{\textit{CNN}}_r $ share the same network parameters. To form the recurrent hidden state $ h^1_i $ of iteration $ i $, the representation $ z_i $ is combined with the hidden state $ h^1_{i-1} $ of the previous iteration:

\begin{equation}
z^1_i = \mbox{\textit{CNN}}_r(x_i)
\end{equation}
\begin{equation}
h^1_i = \phi(W^1_I\ast z^1_i+ W^1_R\ast h^1_{i-1} + b^1)
\end{equation}
where $ W^1_I $ is the convolution filters for input-to-hidden connections, $ W^1_R $ is the convolution filters for hiddent-to-hidden connections between any two consecutive iterations, $ b^1 $ is a bias term. As in RNN, the hidden-to-hidden connections allow contextual information gathered at previous iterations to be passed to the future iterations. Since RACDNN is attentional, the already attended sub-regions can help to guide saliency refinement for the upcoming sub-regions. This is beneficial for the task of saliency detection, as the saliency of an object is highly dependable on its surrounding regions. Different from conventional RNNs that use matrix product (fully-connected network layers) for both input-to-hidden and hidden-to-hidden connections, these connections in our method are convolution operations (convolutional layers) as in \cite{pinheiro2014recurrent}. By using recurrent connections that are convolutional, we can preserve the spatial information of hidden representation $ h^1_i $. As mentioned in Section 3.1, preserving the spatial information of hidden representation between CNN and DecNN is favorable for DecNN's upsizing-related operations.

After obtaining $ h^1_i $, we can then perform saliency refinement on initial saliency maps using $ \mbox{\textit{DecNN}}_r $. The initial saliency maps are generated by the global CNN-DecNN in single forward passes. Instead of replacing the values of initial saliency map with the output of RACDNN at each iteration, the initial saliency map $ r_0 $ is refined cumulatively for $ N $ number of iterations. At iteration $ i $, the saliency map $ r_i $ is refined as

\begin{equation} 
r_i = r_{i-1} + ST(\mbox{\textit{DecNN}}_r(h^1_{i-1}),\tau^{-1}_i)
\end{equation}
Before being added to $ r_i $, the saliency output of $ \mbox{\textit{DecNN}}_r $ is spatially transformed back to the attended sub-region using inverse spatial transformer ($ ST $). For the unattended regions, the saliency refinement values are set as zero and thus those regions do not affect $ r_i $. After $ N $ number of iterations, as in Section 3.1, sigmoid activation function $ \sigma(\cdot) $ is applied to $ r_N $, resulting in the final saliency map $ \bar{S}_r $.

Besides saliency refinement outputs, at every iteration, RACDNN should generate $ \tau $ to determine which sub-region to attend to in the next iteration. A simple way to achieve that is by simply treating $ h^1_i $ as input to a fully-connected network-based regressor. However, to model the sequential dependencies between attended locations, such a simplistic approach is insufficient. This is because $ h^1_i $ should focus mainly on modeling contextual dependencies for saliency refinement, not multiple kinds of dependency. To better model locational dependencies, we propose to add another recurrent layer to RACDNN. The hidden state of the second recurrent layer at iteration $ i $ is denoted by $ h^2_i $ and it is formulated as

\begin{equation}
h^2_i = \phi(W^2_I h^1_i+ W^2_R h^2_{i-1} + b^2)
\end{equation}
where the weights $ W^2_I, W^2_R $ and bias $ b^2 $ are semantically the same as their counterparts in the first recurrent layer in Equation (5). The input of the second recurrent layer is the output of the first recurrent layer, making the RACDNN a stacked recurrent network. Considering the nature of the regression task, we use only fully-connected layers for both recurrent input and hidden connections in the second recurrent layer. Finally, given $ h^2_i $, a $ f_{loc}(\cdot) $ can be used to regress the transformation matrix for the next iteration $ i+1 $:

\begin{equation} 
\tau_{i+1} = f_{loc}(h^2_i) = \phi(W_{loc^2}\phi(W_{loc^1}h^2_i))
\end{equation}
$ W_{loc^1} $ and $ W_{loc^2} $ are respectively the weight matrices of the first and second layers of the two-layered fully-connected network $ f_{loc}(\cdot) $ used in our work.

In RACDNN, the hidden representations $ (h^1_0, h^2_0) $ at the $0$-th iteration are provided by a CNN (sharing the same architectural properties as $ \mbox{\textit{CNN}}_r $) which accepts the whole image region as input. Observing the full image region at the $0$-th iteration helps RACDNN to better decide which sub-regions to attend subsequently.

Similar to the CNN-DecNN used for saliency detection, the loss function of RADCNN is the binary cross-entropy between the final saliency output $ \bar{S}_r $ and the groundtruth saliency map $ \bar{G} $. Since every component in RADCNN is differentiable, errors can be backpropagated to all network layers and parameters of RADCNN, making it trainable with any gradient-based optimization methods (e.g., gradient descent). $ W^1_I $, $ W^1_R $, $ b^1 $, $ W^2_I $, $ W^2_R $, $ b^2 $, $ W_{loc^1} $, $ W_{loc^2} $, and the network weights in $ \mbox{\textit{CNN}}_r $ and $ \mbox{\textit{DecNN}}_r $ are learnable parameters in RADCNN.

\begin{table*}[ht]
	\fontsize{3.2}{3.2}\selectfont
	\begin{center}
		\resizebox{\textwidth}{!}{
			\setlength{\arrayrulewidth}{0.2pt}
			\begin{tabular}{|c||c|c||c|c||c|c||c|c||c|c|}
				\hline
				& \multicolumn{2}{c||}{\textbf{MSRA10K}} & \multicolumn{2}{c||}{\textbf{THUR15K}} & \multicolumn{2}{c||}{\textbf{HKUIS}} & \multicolumn{2}{c||}{\textbf{ECSSD}} & \multicolumn{2}{c|}{\textbf{SED2}} \\ \hline
				in \% & \textbf{F-M} & \textbf{MAE} & \textbf{F-M} & \textbf{MAE} & \textbf{F-M} & \textbf{MAE} & \textbf{F-M} & \textbf{MAE} & \textbf{F-M} & \textbf{MAE} \\ \hline
				\textbf{CNN-DecNN} & 87.91 & 7.03 & 69.28 & 10.42 & 82.48 & 8.10 & 85.72 & 8.72 & 82.79 & \color{red}{9.29} \\ \hline
				\textbf{+ NRACDNN} & 88.62 & 6.85 & 70.39 & 10.46 & 83.74 & 7.88 & 86.65 & 8.43 & 83.99 & 9.30 \\ \hline
				\textbf{+ RACDNN} & \color{red}{89.98} & \color{red}{6.02} & \color{red}{71.12} & \color{red}{9.04} & \color{red}{85.57} & \color{red}{7.03} & \color{red}{87.81} & \color{red}{8.12} & \color{red}{85.35} & \color{red}{9.29} \\ \hline
			\end{tabular}
		}
	\end{center}
	\caption{F-measure scores (\textbf{F-M}) and Mean Absolute Errors (\textbf{MAE}) (compared with baseline methods)}
	\label{table:baseline}
\end{table*}

\section{Implementation Details}
For initial saliency detection, we use a CNN-DecNN independent from the CNN-DecNN used in the saliency refinement stage. The CNN part is initialized from the weights of VGG-CNN-S \cite{chatfield2014return}, a relatively powerful CNN model pre-trained on ImageNet dataset. VGG-CNN-S consists of 5 convolutional layers and 3 fully-connected layers. We discard the fully-connected layers of VGG-CNN-S and retain only its convolutional and pooling layers for network initialization. The CNN accepts $ 224 \times 224 $ RGB images as inputs, and it outputs a $7\times7$ feature maps with 256 feature channels. The DecNN part of the initial CNN-DecNN is a network with 3 convolutional layers ($ 5 \times 5$ kernel size, $1 \times 1 $ stride, $ 2 \times 2 $ zero paddings), and there is an unpooling layer before each convolutional layer. To increase the representational capability of the DecNN without adding too many weight parameters, we append a layer convolution layer with $1 \times 1$ convolution kernel, to each DecNN convolutional layer. At the end of the initial CNN-DecNN, the DecNN outputs a $ 56 \times 56 $ saliency map. The output size of $ 56 \times 56 $ achieves a good balance between computational complexity and saliency pixels details. For performance evaluation, the $ 56 \times 56 $ saliency map is resized to the input image's original size. The initial CNN-DecNN is trained with Adam \cite{kingma2014adam}\underline{} in default learning settings.

As mentioned previously, the $ \mbox{\textit{CNN}}_r $ and $ \mbox{\textit{DecNN}}_r $ used in RACDNN are trained and executed independently of those in the initial CNN-DecNN. On the other hand, $ \mbox{\textit{DecNN}}_r $ is initialized using the pre-trained weights of DecNN of the initial CNN-DecNN. In the recurrent layers of RACDNN, rectified linear unit (ReLU) is employed as the non-linear activation $ \phi(\cdot) $. The feature maps of the hidden state $ h^1_i $ (the first recurrent layer of RACDNN) is of size $ 7 \times 7 $ and has $ 256 $ feature channels. For the second recurrent layer's hidden state $ h^2_i $, the feature representation is a $ 512 $-dimensional vector. The weight parameters $ W_{loc^1} $ and $ W_{loc^2} $ of $ f_{loc}(\cdot) $ are $512\times256$ and $256\times3$ matrices respectively. The number of recurrent iterations of RACDNN (inclusive of the $0$-th iteration) is set to $ 9 $ for all saliency detection experiments. RACDNN is trained using RMSProp \cite{tieleman2012} with an initial learning rate of $ 0.0001 $. The learning rate is reduced by an order of magnitude whenever validation performance stops improving. During training, gradients are hard-clipped to be within the range of $ [-5,5] $ as a way to mitigate the gradient explosion problem which occurs when training recurrent-based networks. To speed up training and improve training convergence, we apply Batch Normalization \cite{ioffe2015batch} to all weight layers (except for recurrent hidden-to-hidden connections) in both the initial CNN-DeCNN and RADCNN.

Most of the saliency detection methods employ object segmentation techniques which can output image segments with consistent saliency values within each segment. Furthermore, the edges of the output segments are sharp. To achieve similar effects, we apply a mean shift-based segmentation method \cite{frintrop2015traditional,garcia2015saliency} to the outputs of RACDNN as a post-processing step.

\section{Saliency Training Datasets}
Learning-based methods require a big amount of training samples to generalize to new examples well. However, most of the saliency detection datasets are too small. It is not possible to train the deep models well if the experimental evaluations are done in such a way that each dataset is split into training, testing and validation sets in proportions. Here, we follow the dataset procedure in one recent deep learning-based saliency detection work \cite{zhao2015saliency}. We train the deep models (initial CNN-DecNN and RADCNN) in our proposed method on saliency datasets different from the datasets used for experimental evaluations. The training datasets we use are: DUT-OMRON \cite{yang2013saliency}, NJU2000 \cite{ju2015depth}, RGBD Salient Object Detection dataset \cite{peng2014rgbd}, and ImageNet segmentation dataset \cite{guillaumin2014imagenet}. The data samples in these datasets reach a total number of 12,430, which is roughly the size of the dataset (with 10,000 samples) used in \cite{zhao2015saliency}. We randomly split the combined datasets into 10,565 training samples and 1865 validation samples. Although the training set is considered large in saliency detection context, it is still small for deep learning methods, and may cause overfitting. Thus, we apply data augmentation in the form of cropping, translation, and color jittering on the training samples.

\section{Experiments}

\begin{table*}[ht]
	\fontsize{3.2}{3.2}\selectfont
	\begin{center}
		\resizebox{\textwidth}{!}{
			\setlength{\arrayrulewidth}{0.2pt}
			\begin{tabular}{|c||c|c||c|c||c|c||c|c||c|c|}
				\hline
				& \multicolumn{2}{c||}{\textbf{MSRA10K}} & \multicolumn{2}{c||}{\textbf{THUR15K}} & \multicolumn{2}{c||}{\textbf{HKUIS}} & \multicolumn{2}{c||}{\textbf{ECSSD}} & \multicolumn{2}{c|}{\textbf{SED2}} \\ \hline
				in \% & \textbf{F-M} & \textbf{MAE} & \textbf{F-M} & \textbf{MAE} & \textbf{F-M} & \textbf{MAE} & \textbf{F-M} & \textbf{MAE} & \textbf{F-M} & \textbf{MAE} \\ \hline
				\textbf{RRWR} \cite{li2015robust} & 84.92 & 12.36 & 59.99 & 17.77 & 71.28 & 17.18 & 74.70 & 18.51 & 77.98 & 16.08 \\\hline
				\textbf{BSCA} \cite{qin2015saliency} & 85.88 & 12.52 & 60.94 & 18.24 & 71.89 & 17.48 & 76.03 & 18.32 & 78.25 & 15.79 \\\hline
				\textbf{DRFI} \cite{jiang2013salient} & 88.07 & 11.82 & 67.02 & 15.03 & 77.31 & 13.45 & 78.70 & 16.59 & 83.86 & 12.70 \\\hline
				\textbf{RBD} \cite{zhu2014saliency} & 85.59 & 10.80 & 59.62 & 15.04 & 72.29 & 14.24 & 71.79 & 17.33 & 82.96 & 12.97 \\\hline
				\textbf{DSR} \cite{li2014saliency} & 83.46 & 12.07 & 61.07 & 14.19 & 73.47 & 14.22 & 73.69 & 17.29 & 78.90 & 14.01 \\\hline
				\textbf{MC} \cite{jiang2013saliency} & 84.76 & 14.51 & 60.96 & 18.38 & 72.34 & 18.40 & 74.18 & 20.37 & 77.10 & 17.96 \\\hline
				\textbf{HS} \cite{shi2015hierarchical} & 84.49 & 14.86 & 58.54 & 21.78 & 70.76 & 21.50 & 73.04 & 22.83 & 80.37 & 11.18 \\\hline
				\textbf{Ours} & \color{red}{89.98} & \color{red}{6.02} & \color{red}{71.12} & \color{red}{9.04} & \color{red}{85.57} & \color{red}{7.03} & \color{red}{87.81} & \color{red}{8.12} & \color{red}{85.35} & \color{red}{9.29} \\\hline
			\end{tabular}
		}
	\end{center}
	\caption{F-measure scores (\textbf{F-M}) and Mean Absolute Errors (\textbf{MAE}) (compared with state-of-the-art methods)}
	\label{table:sota}
\end{table*}

\begin{figure*}[ht]
	\centering
	\includegraphics[width=\linewidth]{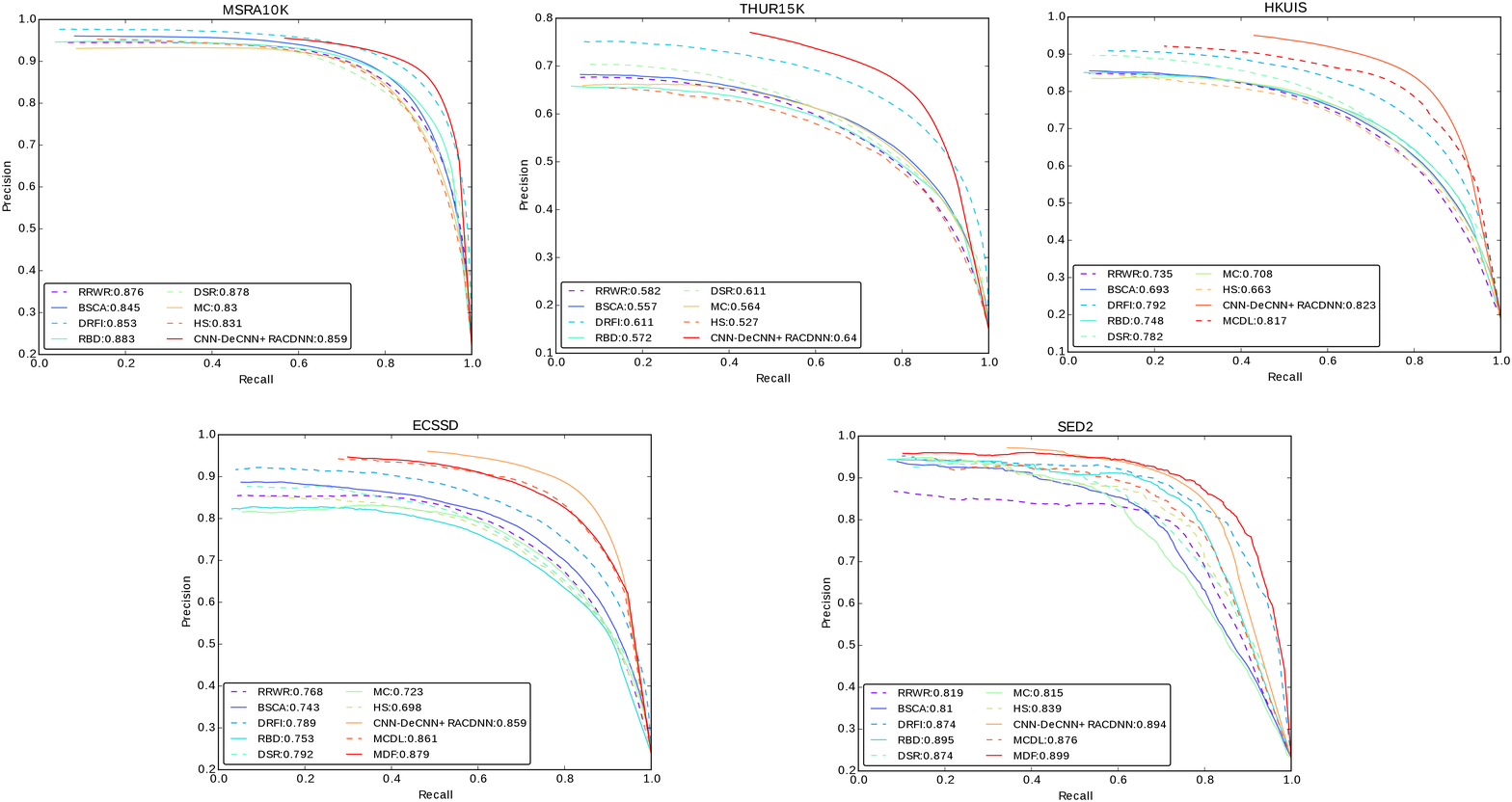}
	\caption{Precision-recall curves, with average precisions}
	\label{fig:prcurve_sota}
\end{figure*}

\subsection{Datasets and Evaluation Metrics}
We evaluate our proposed on a number of challenging saliency detection datasets: \textbf{MSRA10K} \cite{cheng2015global} is by far the largest publicly available saliency detection dataset, containing 10,000 annonated saliency images. \textbf{THUR15K} \cite{cheng2014salientshape} has 6,232 images which belong to five object classes of ``butterfly", ``coffee mug", ``dog jump", ``giraffe", and ``plane". It is challenging because some of its images do not contain any salient object. \textbf{HKUIS} \cite{li2015visual} is a recently released saliency detection dataset with 4,447 annonated images. \textbf{ECSSD} \cite{shi2015hierarchical} is a challenging saliency detection dataset with many semantically meaningful but structurally complex images. It contains 1,000 images. \textbf{SED2} \cite{alpert2007image} is a small saliency dataset having only 100 images. For each image, there are two salient objects.

We evaluate the proposed method based on precision-recall curves, which is the most commonly used evaluation metric for saliency detection. The saliency output is thresholded at integer values within the range of $ [0,255] $. At each threshold value, the binarized saliency output is compared to the binary groundtruth mask to obtain a pair of precision-recall values. Another popular evaluation metric for saliency detection is F-measure, which is a combination of precision and recall values. Following the recent saliency detection benchmark paper \cite{borji2015salient}, we use a weighted F-measure $ F_\beta $ that favors precision more than recall: $ \frac{(1+\beta_2) \mathrm{Precision} \times \mathrm{Recall}}{\beta_2 \mathrm{Precision} + \mathrm{Recall}}
$, where $ \beta_2 $ is set as $ 0.3 $. The reported $ F_\beta $ is the maximum F-measure computed from all precision-recall pairs, which is a good summary of detection performance according to \cite{borji2015salient}.

Even though F-measure is the most commonly used evaluation metric for saliency detection, it is not comprehensive enough as it does not consider true negative saliency labeling. To have a more comprehensive experimental evaluation, we consider another evaluation metric known as Mean Absolute Error (MAE) adopted by \cite{borji2015salient}. MAE is given by: $ \frac{1}{\mathcal{W} \times \mathcal{H}} \sum\limits_{\mathit{n}=1}^\mathcal{W} \sum\limits_{\mathit{m}=1}^\mathcal{H} |\bar{S}(n,m)-\bar{G}(n,m)| $, where $ \mathcal{W} $ and $ \mathcal{H} $ are width and height of saliency map; $ \bar{S} $ is the real-valued  saliency map output normalized to the range of $[0,1]$, and $ \bar{G} $ is the saliency groundtruth. Saliency map binarization is not needed in MAE as it measures the mean of absolute differences between groundtruth saliency pixels and given saliency pixels.

\subsection{Comparison with Baseline Methods}

To highlight the advantages of recurrent attention mechanism in the proposed network RACDNN, we use CNN-DecNN as one of the baseline methods in our experiments. Compared to the proposed method, the baseline CNN-DecNN has no recurrent attention mechanism to perform iterative saliency refinement. The other baseline method is a CNN-DecNN paired with a non-recurrent attentional convolutional-deconvolutional network (NACDNN) in place of RACDNN. NACDNN is a RACDNN variant whose layers $ h^1 $ and $ h^2 $ are made non-recurrent. By removing the recurrent connections, NACDNN cannot learn context-aware features useful for saliency refinement despite having attention mechanism. At each computational iteration, NACDNN works almost like a CNN-DeCNN except that it has a localization network $ f_{loc} (\cdot) $ that accepts CNN's output as input and outputs spatial transformation matrix.

To compare the proposed method with baseline methods, we use F-measure and MAE as evaluation metrics. The F-measure scores and Mean Square Errors (MAEs) for comparisons with the baselines are shown in Table \ref{table:baseline}. On all of the five datasets and two evaluation metrics, the proposed method achieves better results than both the baseline methods. This shows that the RACDNN can help to improve the saliency map outputs of CNN-DecNN, using a recurrent attention mechanism to alleviate the scale issues of CNN-DecNN, and to learn region-based contextual dependencies not easily modeled by mere convolutional and deconvolutional network operations. The second baseline method NRACDNN that has attention mechanism performs better than the non-attentional first baseline. However, due to the lack of recurrent connections, NRACDNN is inferior to RACDNN because it does not exploit contextual information from past iterations for saliency refinement.

\begin{figure*}[ht]
	\includegraphics[width=\linewidth]{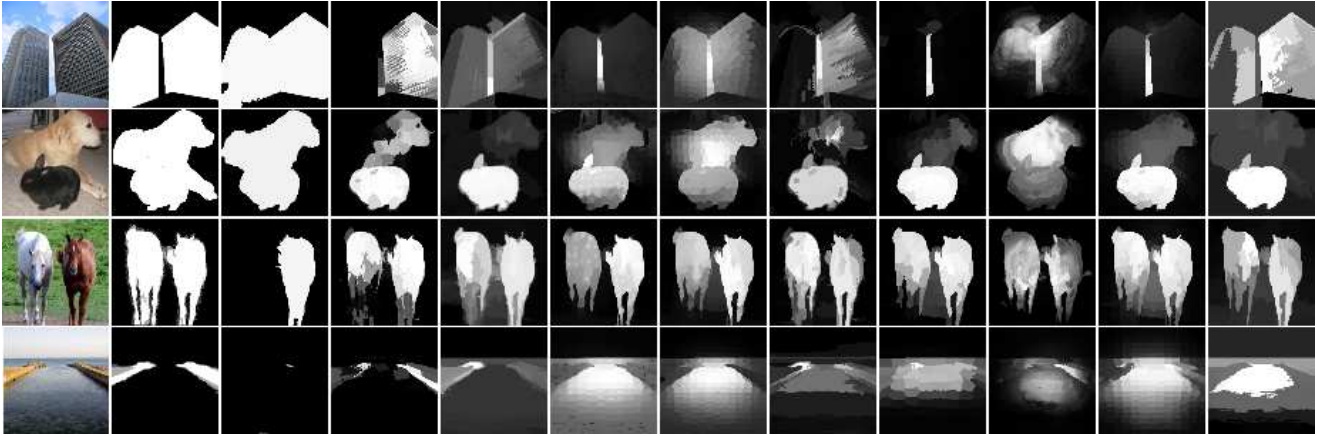}
	\caption{Qualitative saliency results of some evaluated images. From the leftmost column: input image, saliency groundtruth, the saliency output maps of our proposed method (CNN-DecNN + RACDNN) with mean-shift post-processing, MCDL \cite{zhao2015saliency}, MDF \cite{li2015visual}, RRWR \cite{li2015robust}, BSCA \cite{qin2015saliency}, DRFI \cite{jiang2013salient}, RBD \cite{zhu2014saliency}, DSR \cite{li2014saliency}, MC \cite{jiang2013saliency}, and HS \cite{shi2015hierarchical}.}
	\label{fig:back_qualitative}
\end{figure*}

\subsection{Comparison with State-of-the-art Methods}
In addition to the baseline methods, we compare the proposed method ``CNN-DecNN + RACDNN" with several state-of-the-art saliency detection methods: RRWR \cite{li2015robust}, BSCA \cite{qin2015saliency}, DRFI \cite{jiang2013salient}, RBD \cite{zhu2014saliency}, DSR \cite{li2014saliency}, MC \cite{jiang2013saliency}, and HS \cite{shi2015hierarchical}. DRFI, RBD, DSR, MC, and HS are the top-performing methods evaluated in \cite{borji2015salient}, while RRWR and BSCA are two very recent saliency detection works. To obtain the results for these methods, we run the original codes provided by the authors with recommended parameter settings. The precision-recall curves are given in Figure \ref{fig:prcurve_sota}. We compute the curves based on the saliency maps generated by the proposed method. In overall, the proposed method ``CNN-DecNN + RACDNN" performs better than the evaluated state-of-the-art methods. Especially in datasets with complex scenes (ECSSD \& HKUIS), the performance gains of the proposed method over the state-of-the-art methods are more noticeable.

We also compare the proposed method ``CNN-DecNN + RACDNN" with the state-of-the-art methods in terms of F-measure scores and Mean Square Errors (MAEs) (Table \ref{table:sota}). In these evaluation metrics, its performance gains over the other methods are very significant. For the HKUIS and ECSSD dataset, the F-measure improvements of the proposed method over the next top-performing method DRFI are more than 5\%. The proposed method also pushes down the MAEs on these challenging datasets by a large margin. 

Besides quantitative results, we show some qualitative results in Figure \ref{fig:back_qualitative}. The proposed method ``CNN-DecNN + RACDNN" can better detect multiple intermingled salient objects, as shown in the second image with a dog and a rabbit. Our method is the only one that can detect both objects well. The success of our method on this image is attributed to the attention mechanism that allows it to attend to different object regions for local refinement, making it is less likely to be negatively affected by distant noises and other objects. However, the proposed method tends to fail to detect salient objects which are mostly made up of background-like colors and textures (e.g., sky: third image, soil: fourth image).

To further evaluate the proposed method ``CNN-DecNN + RACDNN", we compare it with two recent deep learning-based saliency detection methods (MCDL \cite{zhao2015saliency} and MDF \cite{li2015visual}) on HKUIS, ECSSD, and SED2 datasets. We use the trained models provided by the authors. The F-measure scores and MAEs are given in Table \ref{table:deep}, showing that the proposed method is comparable to both MCDL and MDF in terms of F-measure, but outperforming them in terms of MAEs.

\begin{table}[h]
	\begin{center}
		\resizebox{\columnwidth}{!}{
			\begin{tabular}{|c||c|c||c|c||c|c|}
				\hline
				 & \multicolumn{2}{c||}{\textbf{HKUIS}} & \multicolumn{2}{c||}{\textbf{ECSSD}} & \multicolumn{2}{c|}{\textbf{SED2}} \\ \hline
				in \% & \textbf{F-M} & \textbf{MAE} & \textbf{F-M} & \textbf{MAE} & \textbf{F-M} & \textbf{MAE} \\ \hline
				\textbf{MCDL} \cite{zhao2015saliency} & 80.85 & 9.13 & 83.74 & 10.20 & 81.37 & 11.45 \\ \hline
				\textbf{MDF} \cite{li2015visual} & 86.01* & 12.93* & 83.06 & 10.81 
				& \color{red}{86.23} & 11.18 \\ \hline
				\textbf{Ours}  & \color{red}{85.57} & \color{red}{7.03} & \color{red}{87.81} & \color{red}{8.12} & 85.35 & \color{red}{9.29} \\ \hline
			\end{tabular}
		}
	\end{center}
	\caption{Comparison with deep learning-based methods. *MDF is trained on a subset of HKUIS, and then evaluated on the remaining HKUIS samples.}
	\label{table:deep}
\end{table}

\section{Conclusion}
In this paper, we introduce a novel method of using recurrent attention and convolutional-deconvolutional network to tackle the saliency detection problem. The proposed method has shown to be very effective experimentally. Still, the performance of proposed method may be limited by the quality of the initial saliency maps. To overcome such limitation, the recurrent attentional network can be potentially revamped to detect saliency from scratch in end-to-end manner. Also, this work can be readily adapted for other vision tasks that require pixel-wise prediction \cite{dosovitskiy2015learning,long2015fully}. \\

\noindent\textbf{Acknowledgement:} The research is supported by Singapore Ministry of Education (MOE) Tier 2 ARC28/14, and Singapore A*STAR Science and Engineering Research Council PSF1321202099.

{\small
\bibliographystyle{ieee}
\bibliography{draftbib}

\begin{thebibliography}{10}\itemsep=-1pt

\bibitem{achanta2009frequency}
R.~Achanta, S.~Hemami, F.~Estrada, and S.~Susstrunk.
\newblock Frequency-tuned salient region detection.
\newblock In {\em IEEE Conference on Computer Vision and Pattern Recognition},
  pages 1597--1604. IEEE, 2009.

\bibitem{alpert2007image}
S.~Alpert, M.~Galun, R.~Basri, and A.~Brandt.
\newblock Image segmentation by probabilistic bottom-up aggregation and cue
  integration.
\newblock In {\em IEEE Conference on Computer Vision and Pattern Recognition},
  pages 1--8, 2007.

\bibitem{ba2015multiple}
J.~Ba, V.~Mnih, and K.~Kavukcuoglu.
\newblock Multiple object recognition with visual attention.
\newblock In {\em International Conference on Learning Representations}, 2015.

\bibitem{borji2015salient}
A.~Borji, M.~Cheng, H.~Jiang, and J.~Li.
\newblock Salient object detection: A benchmark.
\newblock {\em IEEE Transactions on Image Processing}, 24(12):5706--5722, 2015.

\bibitem{borji2012adaptive}
A.~Borji, S.~Frintrop, D.~N. Sihite, and L.~Itti.
\newblock Adaptive object tracking by learning background context.
\newblock In {\em IEEE Computer Society Conference on Computer Vision and
  Pattern Recognition Workshops}, pages 23--30. IEEE, 2012.

\bibitem{chatfield2014return}
K.~Chatfield, K.~Simonyan, A.~Vedaldi, and A.~Zisserman.
\newblock Return of the devil in the details: Delving deep into convolutional
  nets.
\newblock In {\em British Machine Vision Conference}, 2014.

\bibitem{chen2009sketch2photo}
T.~Chen, M.-M. Cheng, P.~Tan, A.~Shamir, and S.-M. Hu.
\newblock Sketch2photo: internet image montage.
\newblock {\em ACM Transactions on Graphics}, 28(5):124, 2009.

\bibitem{cheng2014salientshape}
M.-M. Cheng, N.~Mitra, X.~Huang, and S.-M. Hu.
\newblock Salientshape: group saliency in image collections.
\newblock {\em The Visual Computer}, 30(4):443--453, 2014.

\bibitem{cheng2015global}
M.-M. Cheng, N.~J. Mitra, X.~Huang, P.~H.~S. Torr, and S.-M. Hu.
\newblock Global contrast based salient region detection.
\newblock {\em IEEE Transactions on Pattern Analysis and Machine Intelligence},
  37(3):569--582, 2015.

\bibitem{dosovitskiy2015learning}
A.~Dosovitskiy, J.~Tobias~Springenberg, and T.~Brox.
\newblock Learning to generate chairs with convolutional neural networks.
\newblock In {\em IEEE Conference on Computer Vision and Pattern Recognition},
  pages 1538--1546, 2015.

\bibitem{elman1990finding}
J.~L. Elman.
\newblock Finding structure in time.
\newblock {\em Cognitive science}, 14(2):179--211, 1990.

\bibitem{frintrop2015traditional}
S.~Frintrop, T.~Werner, and G.~Martin~Garcia.
\newblock Traditional saliency reloaded: A good old model in new shape.
\newblock In {\em IEEE Conference on Computer Vision and Pattern Recognition},
  June 2015.

\bibitem{garcia2015saliency}
G.~Garcia, E.~Potapova, T.~Werner, M.~Zillich, M.~Vincze, and S.~Frintrop.
\newblock Saliency-based object discovery on rgb-d data with a late-fusion
  approach.
\newblock In {\em IEEE International Conference on Robotics and Automation},
  pages 1866--1873, 2015.

\bibitem{gregor2015draw}
K.~Gregor, I.~Danihelka, A.~Graves, D.~Rezende, and D.~Wierstra.
\newblock Draw: A recurrent neural network for image generation.
\newblock In {\em International Conference on Machine Learning}. JMLR Workshop
  and Conference Proceedings, 2015.

\bibitem{gu2015recent}
J.~Gu, Z.~Wang, J.~Kuen, L.~Ma, A.~Shahroudy, B.~Shuai, T.~Liu, X.~Wang, and
  G.~Wang.
\newblock Recent advances in convolutional neural networks.
\newblock {\em arXiv preprint arXiv:1512.07108}, 2015.

\bibitem{guillaumin2014imagenet}
M.~Guillaumin, D.~K{\"u}ttel, and V.~Ferrari.
\newblock Imagenet auto-annotation with segmentation propagation.
\newblock {\em International Journal of Computer Vision}, 110(3):328--348,
  2014.

\bibitem{harel2006graph}
J.~Harel, C.~Koch, and P.~Perona.
\newblock Graph-based visual saliency.
\newblock In {\em Advances in Neural Information Processing Systems}, pages
  545--552, 2006.

\bibitem{he2015delving}
K.~He, X.~Zhang, S.~Ren, and J.~Sun.
\newblock Delving deep into rectifiers: Surpassing human-level performance on
  imagenet classification.
\newblock In {\em IEEE International Conference on Computer Vision}, 2015.

\bibitem{hou2007saliency}
X.~Hou and L.~Zhang.
\newblock Saliency detection: A spectral residual approach.
\newblock In {\em IEEE Conference on Computer Vision and Pattern Recognition},
  pages 1--8. IEEE, 2007.

\bibitem{ioffe2015batch}
S.~Ioffe and C.~Szegedy.
\newblock Batch normalization: Accelerating deep network training by reducing
  internal covariate shift.
\newblock {\em Journal of Machine Learning Research}, 2015.

\bibitem{itti1998model}
L.~Itti, C.~Koch, and E.~Niebur.
\newblock A model of saliency-based visual attention for rapid scene analysis.
\newblock {\em IEEE Transactions on Pattern Analysis and Machine Intelligence},
  pages 1254--1259, 1998.

\bibitem{jaderberg2015spatial}
M.~Jaderberg, K.~Simonyan, A.~Zisserman, and K.~Kavukcuoglu.
\newblock Spatial transformer networks.
\newblock In {\em Advances in Neural Information Processing Systems}, 2015.

\bibitem{jiang2013saliency}
B.~Jiang, L.~Zhang, H.~Lu, C.~Yang, and M.-H. Yang.
\newblock Saliency detection via absorbing markov chain.
\newblock In {\em IEEE International Conference on Computer Vision}, ICCV '13,
  pages 1665--1672, 2013.

\bibitem{jiang2013salient}
H.~Jiang, J.~Wang, Z.~Yuan, Y.~Wu, N.~Zheng, and S.~Li.
\newblock Salient object detection: A discriminative regional feature
  integration approach.
\newblock In {\em IEEE Conference on Computer Vision and Pattern Recognition},
  pages 2083--2090. IEEE, 2013.

\bibitem{ju2015depth}
R.~Ju, Y.~Liu, T.~Ren, L.~Ge, and G.~Wu.
\newblock Depth-aware salient object detection using anisotropic
  center-surround difference.
\newblock {\em Signal Processing: Image Communication}, 2015.

\bibitem{judd2009learning}
T.~Judd, K.~Ehinger, F.~Durand, and A.~Torralba.
\newblock Learning to predict where humans look.
\newblock In {\em IEEE Conference on Computer Vision and Pattern Recognition},
  pages 2106--2113. IEEE, 2009.

\bibitem{kingma2014adam}
D.~Kingma and J.~Ba.
\newblock Adam: A method for stochastic optimization.
\newblock In {\em International Conference on Learning Representations}, 2014.

\bibitem{li2015robust}
C.~Li, Y.~Yuan, W.~Cai, Y.~Xia, and D.~Dagan~Feng.
\newblock Robust saliency detection via regularized random walks ranking.
\newblock In {\em IEEE Conference on Computer Vision and Pattern Recognition},
  June 2015.

\bibitem{li2015visual}
G.~Li and Y.~Yu.
\newblock Visual saliency based on multiscale deep features.
\newblock In {\em IEEE Conference on Computer Vision and Pattern Recognition},
  2015.

\bibitem{li2014saliency}
X.~Li, H.~Lu, L.~Zhang, X.~Ruan, and M.-H. Yang.
\newblock Saliency detection via dense and sparse reconstruction.
\newblock In {\em IEEE International Conference on Computer Vision}, ICCV '13,
  pages 2976--2983, 2013.

\bibitem{nian2015predicting}
N.~Liu, J.~Han, D.~Zhang, S.~Wen, and T.~Liu.
\newblock Predicting eye fixations using convolutional neural networks.
\newblock In {\em IEEE Conference on Computer Vision and Pattern Recognition},
  pages 362--370, June 2015.

\bibitem{liu2011learning}
T.~Liu, Z.~Yuan, J.~Sun, J.~Wang, N.~Zheng, X.~Tang, and H.-Y. Shum.
\newblock Learning to detect a salient object.
\newblock {\em IEEE Transactions on Pattern Analysis and Machine Intelligence},
  33(2):353--367, 2011.

\bibitem{long2015fully}
J.~Long, E.~Shelhamer, and T.~Darrell.
\newblock Fully convolutional networks for semantic segmentation.
\newblock In {\em IEEE Conference on Computer Vision and Pattern Recognition},
  pages 3431--3440, 2015.

\bibitem{margolin2013what}
R.~Margolin, A.~Tal, and L.~Zelnik-Manor.
\newblock What makes a patch distinct?
\newblock In {\em IEEE Conference on Computer Vision and Pattern Recognition},
  pages 1139--1146, 2013.

\bibitem{mnih2014recurrent}
V.~Mnih, N.~Heess, A.~Graves, et~al.
\newblock Recurrent models of visual attention.
\newblock In {\em Advances in Neural Information Processing Systems}, pages
  2204--2212, 2014.

\bibitem{noh2015learning}
H.~Noh, S.~Hong, and B.~Han.
\newblock Learning deconvolution network for semantic segmentation.
\newblock In {\em IEEE International Conference on Computer Vision}, 2015.

\bibitem{peng2014rgbd}
H.~Peng, B.~Li, W.~Xiong, W.~Hu, and R.~Ji.
\newblock Rgbd salient object detection: a benchmark and algorithms.
\newblock In {\em European Conference on Computer Vision}, pages 92--109, 2014.

\bibitem{pinheiro2014recurrent}
P.~Pinheiro and R.~Collobert.
\newblock Recurrent convolutional neural networks for scene labeling.
\newblock In {\em International Conference on Machine Learning}, pages 82--90,
  2014.

\bibitem{qin2015saliency}
Y.~Qin, H.~Lu, Y.~Xu, and H.~Wang.
\newblock Saliency detection via cellular automata.
\newblock In {\em IEEE Conference on Computer Vision and Pattern Recognition},
  June 2015.

\bibitem{ren2014region}
Z.~Ren, S.~Gao, L.-T. Chia, and I.~W.-H. Tsang.
\newblock Region-based saliency detection and its application in object
  recognition.
\newblock {\em IEEE Transactions Circuits and Systems for Video Technology},
  24(5):769--779, 2014.

\bibitem{shi2015hierarchical}
J.~Shi, Q.~Yan, L.~Xu, and J.~Jia.
\newblock Hierarchical image saliency detection on extended cssd.
\newblock {\em IEEE Transactions on Pattern Analysis and Machine Intelligence},
  PP(99):1--1, 2015.

\bibitem{tieleman2012}
T.~Tieleman and G.~Hinton.
\newblock {Lecture 6.5---RmsProp: Divide the gradient by a running average of
  its recent magnitude}.
\newblock COURSERA: Neural Networks for Machine Learning, 2012.

\bibitem{vig2014large}
E.~Vig, M.~Dorr, and D.~Cox.
\newblock Large-scale optimization of hierarchical features for saliency
  prediction in natural images.
\newblock In {\em IEEE Conference on Computer Vision and Pattern Recognition},
  pages 2798--2805. IEEE, 2014.

\bibitem{wang2015deep}
L.~Wang, H.~Lu, X.~Ruan, and M.-H. Yang.
\newblock Deep networks for saliency detection via local estimation and global
  search.
\newblock In {\em IEEE Conference on Computer Vision and Pattern Recognition},
  June 2015.

\bibitem{yang2013saliency}
C.~Yang, L.~Zhang, H.~Lu, X.~Ruan, and M.-H. Yang.
\newblock Saliency detection via graph-based manifold ranking.
\newblock In {\em IEEE Conference on Computer Vision and Pattern Recognition},
  pages 3166--3173. IEEE, 2013.

\bibitem{yang2012top}
J.~Yang and M.-H. Yang.
\newblock Top-down visual saliency via joint crf and dictionary learning.
\newblock In {\em IEEE Conference on Computer Vision and Pattern Recognition},
  pages 2296--2303. IEEE, 2012.

\bibitem{zhang2015co}
D.~Zhang, J.~Han, C.~Li, and J.~Wang.
\newblock Co-saliency detection via looking deep and wide.
\newblock In {\em IEEE Conference on Computer Vision and Pattern Recognition},
  June 2015.

\bibitem{zhang2008sun}
L.~Zhang, M.~H. Tong, T.~K. Marks, H.~Shan, and G.~W. Cottrell.
\newblock Sun: A bayesian framework for saliency using natural statistics.
\newblock {\em Journal of vision}, 8(7):32, 2008.

\bibitem{zhao2015saliency}
R.~Zhao, W.~Ouyang, H.~Li, and X.~Wang.
\newblock Saliency detection by multi-context deep learning.
\newblock In {\em IEEE Conference on Computer Vision and Pattern Recognition},
  pages 1265--1274, 2015.

\bibitem{zhu2014saliency}
W.~Zhu, S.~Liang, Y.~Wei, and J.~Sun.
\newblock Saliency optimization from robust background detection.
\newblock In {\em IEEE Conference on Computer Vision and Pattern Recognition},
  CVPR '14, pages 2814--2821, 2014.

\end{thebibliography}
}

\end{document}